\pdfoutput=1

\documentclass[11pt]{article}

\usepackage[]{ACL2023}

\usepackage{times}
\usepackage{latexsym}

\usepackage[T1]{fontenc}

\usepackage[utf8]{inputenc}

\usepackage{microtype}

\usepackage{inconsolata}

\usepackage{multirow}
\usepackage{amsmath}
\usepackage{amsfonts}
\usepackage{graphicx}
\usepackage{arydshln}
\usepackage{comment}
\usepackage{booktabs}

\newcommand{\red}[1]{\textcolor{red}{{#1}}}

\newcommand{\modelshort}[1]{\textsc{FineGrainFact}}

\title{Interpretable Automatic Fine-grained Inconsistency Detection in \\Text Summarization}

\author{Hou Pong Chan\textsuperscript{1} ~~~ Qi Zeng\textsuperscript{2}~~~ Heng Ji\textsuperscript{2}\\
\textsuperscript{1}University of Macau ~~~ 
\textsuperscript{2}University of Illinois Urbana-Champaign\\
\texttt{hpchan@um.edu.mo}, \texttt{\{qizeng2, hengji\}@illinois.edu} \\
  \\}

\begin{document}
\maketitle
\begin{abstract}

Existing factual consistency evaluation approaches for text summarization provide binary predictions and limited insights into the weakness of summarization systems. Therefore, we propose the task of fine-grained inconsistency detection, the goal of which is to predict the fine-grained types of factual errors in a summary. Motivated by how humans inspect factual inconsistency in summaries, we propose an interpretable fine-grained inconsistency detection model, \modelshort~, which explicitly represents the facts in the documents and summaries with semantic frames extracted by semantic role labeling, and highlights the related semantic frames to predict inconsistency. The highlighted semantic frames help verify predicted error types and correct inconsistent summaries. Experiment results demonstrate that our model outperforms strong baselines and provides evidence to support or refute the summary.\footnote{Code and data are available at \url{https://github.com/kenchan0226/fineGrainedFact}}

\end{abstract}

\section{Introduction}

Prior work~\cite{DBLP:conf/naacl/QAFactEval22,DBLP:conf/emnlp/DAE20,DBLP:journals/tacl/summac22} formulates the problem of factual inconsistency detection as a binary classification task, which predicts whether a summary is consistent with the source document. 
However, these approaches have two drawbacks. 
First, they cannot predict the types of factual errors made by a summary and thus provide limited insights into the weakness of summarization systems. 
Although  recent studies~\cite{DBLP:conf/naacl/Frank21,DBLP:journals/corr/aggrefact21,DBLP:conf/naacl/AnnotatingFineGrained21} have manually inspected the types of factual errors in summaries, there is \textit{no existing work on automatic detection of fine-grained factual inconsistency. }

Second, existing models typically cannot explain which portions of the document are used to detect the inconsistency in the input summary. 
In order to verify and correct an inconsistent summary, humans still need to read the entire source document to find the supporting evidence. 
\citet{DBLP:conf/emnlp/Factcc20} introduce an auxiliary task to extract the supporting spans in the document  for inconsistency detection, which \textit{requires expensive ground-truth labels of supporting spans}. 

To address the first limitation, we propose the \textbf{fine-grained factual inconsistency detection} task. 
The goal is to predict the types of factual inconsistency in a summary. 
We show examples of different factual error types in Table~\ref{table:error-types-examples}.

\begin{table*}[t]
\small
\centering
\begin{tabular}{p{2\columnwidth}}
\toprule
\textbf{Source text} \\ \hline
Marcy Smith was woken up by her son David to find their house in Glovertown, Newfoundland and Labrador, completely engulfed in flames ... 
Mrs Smith said if it wasn't for her son, she and her daughter probably wouldn't have survived.
\textbf{David was on FaceTime to his father at the time}, so was the only one awake and \textbf{saw the flames out of the corner of his eye} ... \\ \bottomrule
\end{tabular}
\begin{tabular}{p{0.95\columnwidth}p{1\columnwidth}}
\textbf{Error type}                                                                                 & \textbf{Example summary}                                                                                                                                \\ \hline
\textbf{Extrinsic noun phrase error:} Errors that add new object(s), subject(s), or prepositional object(s) that cannot be inferred from the source article.                                                                         & David was using FaceTime with \red{\textit{Maggie Smith}} and saw the flames. \\ \hline
\textbf{Intrinsic noun phrase error:} Errors that misrepresent object(s), subject(s), or prepositional object(s) from the source article.     & David was using FaceTime with \red{\textit{Marcy Smith}} and saw the flames. \\ \hline
\textbf{Extrinsic predicate error:} Errors that add new main verb(s) or adverb(s) that cannot be inferred from the source article.       & David was \red{\textit{eating}} and saw the flames.  \\ \hline
\textbf{Intrinsic predicate error:} Errors that misrepresent main verb(s) or adverb(s) from the source article. 
& David was \red{\textit{engulfed}} and saw the flames. 
\\ \bottomrule
\end{tabular}
\caption{
A text document and example summaries with different factual error types according to the typology defined by \citet{DBLP:journals/corr/aggrefact21}. 
The errors in the sample summaries are in red color and italicized. We bold the text spans from the document that refute the sample summaries. 
}
\vspace{-0.1in}
\label{table:error-types-examples}
\end{table*}

To solve the second challenge, 
we further introduce an \textbf{interpretable fine-grained inconsistency detection model} (\modelshort~) that does not require any label of supporting text spans, inspired by how humans verify the consistency of a summary. 
When humans annotate the factual error types of a summary, they first identify facts in the document that are relevant to the summary and then determine the factual error types in the summary. 
Following this intuition, our model first extracts facts from the document and summary using Semantic Role Labeling (SRL).  
We consider each extracted semantic frame as a fact since a semantic frame captures a predicate and its associated arguments to answer the question of ``who did what to whom''. 
After fact extraction, a document fact attention module enables the classifier to focus on the facts in the document that are most related to the facts in the summary. 
By highlighting the facts in the document with the highest attention scores, our model can explain which facts in the document are most pertinent to inconsistency detection.

Experiment results show that our model outperforms strong baselines
in detecting factual error types. 
Moreover, the document facts highlighted by our model can provide evidence to support or refute the input summary, which can potentially help users to verify the predicted error types and correct an inconsistent summary.

\section{Task Definition}
The goal of the fine-grained inconsistency detection task is to predict the types of factual errors in a summary. 
We frame it as a multi-label classification problem as follows.
Given a pre-defined set of $l$ factual error types $\{e_1, \ldots, e_l \}$, a document $\mathbf{d}$, and a summary $\mathbf{s}$, the goal is to predict a binary vector $\mathbf{y}\in \{0,1\}^{l}$ where each element $y_{i}$ indicates the presence of one type of factual errors.

We follow the typology of factual error types proposed by \cite{DBLP:journals/corr/aggrefact21}, which include
\textit{intrinsic noun phrase error}, \textit{extrinsic noun phrase error}, \textit{intrinsic predicate error}, and \textit{extrinsic predicate error}.  
The definitions and examples of these error types are presented in Table~\ref{table:error-types-examples}.

\section{Our \modelshort~ Model}

\begin{figure}[t]
\centering
\includegraphics[width=\linewidth]{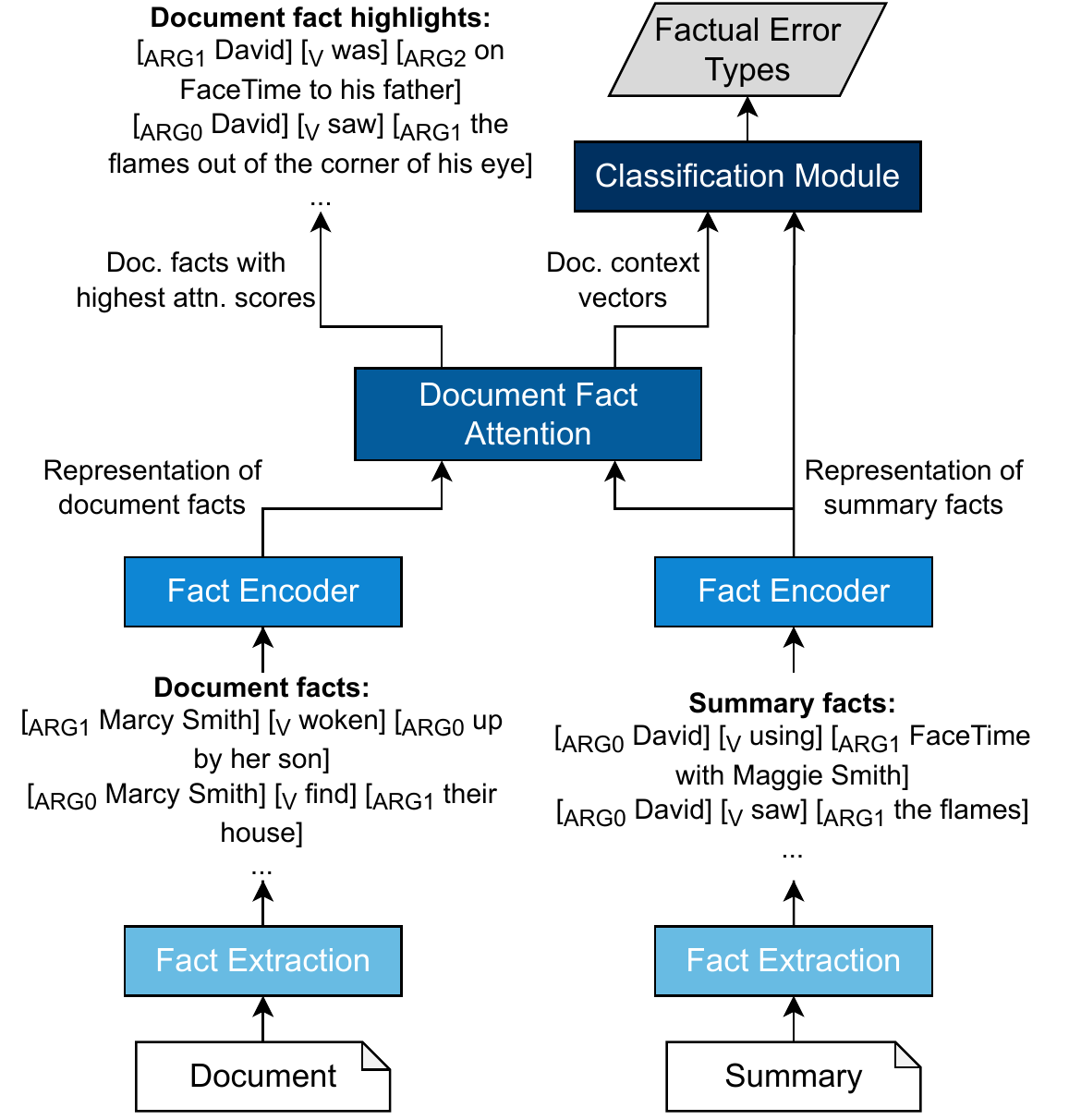}
\vspace{-0.1in}
\caption{
The architecture of \modelshort~. The fact extraction module represents facts from the input document and summary with semantic frames. The document fact attention module queries the document facts with summary facts and highlights those with the highest attention scores.
Based on the retrieved highlighted document context and summary facts, the classification module predicts the factual error types.
}
\label{fig:model}
\vspace{-0.2in}
\end{figure}

The model architecture is illustrated in Figure~\ref{fig:model}.  

\paragraph{Fact extraction. }
To represent facts from the input document and summary, we extract semantic frames with a BERT-based semantic role labeling (SRL) tool~\cite{DBLP:journals/corr/SRL19}.
A semantic frame contains a predicate and its arguments, e.g., $[_{\text{ARG0}}\text{David}][_{\text{V}}\text{saw}][_{\text{ARG1}}\text{the flame}]$.  
We use ${f}^{doc}_{i}$ and ${f}^{sum}_{i}$ to denote the $i$-th fact in the document and summary, respectively. 

\paragraph{Fact encoder. }

We first represent tokens in the concatenated sequence of the input document and summary by fusing hidden states across all layers in Adapter-BERT~\cite{DBLP:conf/icml/AdapterBERT19} with max pooling.
To represent facts, we apply attentive pooling to all tokens in the semantic frame under the assumption that different tokens in a fact should contribute differently to the fact representation. 
Given the token representations $\textbf{t}_{j}$, we calculate the attention scores 
$\alpha_{j} = \exp (\phi(\textbf{t}_{j}))/ \sum_{j=1}^{m} \exp (\phi(\textbf{t}_{j}))$,
and represent each document or summary fact as $\mathbf{f}_{i} = \sum_{j=1}^{m} \alpha_{j} (\phi(\textbf{t}_{j}))$, where $m$ is the number of tokens in the fact and $\phi$ is a two-layer fully-connected network.

\paragraph{Document Fact Attention module. }
This module aims to retrieve the facts in the document that are related to the facts in the summary. 
We first concatenate the document fact representations into a document fact matrix $\mathbf{F}^{doc}$. 
We attend each summary fact ${f}^{sum}_{i}$ to the document fact matrix to compute a \textbf{document context vector}: $\mathbf{c}_{i} = \textsc{MultiHeadAtt}(\mathbf{f}^{sum}_{i}, \mathbf{F}^{doc}, \mathbf{F}^{doc})$, 
where $\mathbf{f}^{sum}_{i}$ acts as the query, $\mathbf{F}^{doc}$ is used as the key and value. 
The document context vector $\mathbf{c}_{i}$ captures the information of the facts in the document that are related to the summary fact ${f}^{sum}_{i}$.

For each document fact, we sum up its attention scores received from all summary facts as its importance score. 
Concretely, we use $\alpha_{j\rightarrow i}$ to denote the sum of attention scores injected from the $j$-th summary fact to the $i$-th document fact over all attention heads. 
The importance score of a document fact ${f}^{doc}_{i}$ is defined as $\sum_{j=1}^{n} \alpha_{j\rightarrow i}$, where $n$ is the total number of facts in the summary. 
Then, we return the top $k$ document facts with the highest importance scores as the \textbf{document fact highlights}, where $k$ is a hyper-parameter.

\paragraph{Classification module.}
A linear classifier predicts the probability of each factual error type based on the concatenation of the representations of summary facts and document context vectors. 
Specifically, we first use mean pooling to fuse all summary fact representation vectors and all document context vectors into two fixed-size vectors:
$\bar{\mathbf{f}}^{sum} = \frac{1}{n} \sum_{i=1}^{n} \mathbf{f}^{sum}_{i}$, $\bar{\mathbf{c}} = \frac{1}{n} \sum_{i=1}^{n} \mathbf{c}_{i}$. 
These two vectors contain the information of all facts in the summary and the information of all document facts that are related to the summary. 
Next, we feed the concatenation of $\bar{\mathbf{f}}^{sum}$ and $\bar{\mathbf{c}}$ to a linear classification layer to predict the probability of each factual error type: $p(\mathbf{y}) = \sigma( \mathbf{W} [ \bar{\mathbf{f}}^{sum}; \bar{\mathbf{c}}] + b )$, where $\mathbf{W} \in \mathbb{R}^{d\times l}, b\in \mathbb{R}$, $d$ is the hidden size of Adapter-BERT, $\sigma$ denotes the sigmoid function.

\paragraph{Training objective.}
We train our model with weighted binary cross-entropy (BCE) loss, The technical details are in Appendix~\ref{appendix:loss-function}.

\section{Experiments}
\subsection{Setup}

\begin{table*}[t]
\centering
\small
\begin{tabular}{l c c c c c c c c c c c c}
\toprule
            & \multicolumn{2}{c}{\textbf{SOTA}} & \multicolumn{2}{c}{\textbf{XFORMER}} & \multicolumn{2}{c}{\textbf{OLD}} & \multicolumn{2}{c}{\textbf{REF}} & \multicolumn{2}{c}{\textbf{All}} \\ 
\textbf{Model}       & \textbf{F1}   & \textbf{BACC}   & \textbf{F1}     & \textbf{BACC}    & \textbf{F1}   & \textbf{BACC} & \textbf{F1}   & \textbf{BACC}   & \textbf{F1}     & \textbf{BACC}   \\ \hline
\textsc{BERT}        &    32.15  & 62.45  & 45.79  & 59.79  & 47.48  & 65.13  & 41.70  & 57.08  & 45.14  & 63.59      \\
\textsc{AdapterBert} & 33.87 & 62.95  & 46.01  & 59.21  & 46.87  & 63.72  & 42.42  & 57.57  & 45.06  & 63.05  \\
\textsc{FactCC-Multi}        &  34.35      &     64.04  &    45.20      &   \textbf{60.28}     &   47.43     &  64.47   &   36.52     &  48.90     &  44.59        &      63.05    \\
\textsc{FactGraph-Multi} &      34.24      &    63.62        &      37.03         &    56.89         &    38.12         &  59.76  &   35.66     &  52.63     &    37.47      &     59.61    \\  
\hdashline
\modelshort~       & \textbf{35.10} & \textbf{64.08} & \textbf{46.02}  & {59.42}  & \textbf{48.63}  & \textbf{65.48}  & \textbf{46.44}  & \textbf{61.81}  & \textbf{46.43}  & \textbf{64.31} \\
\hspace{2mm} $-$ Doc. Fact Attention & 34.77 & 63.12  & 45.61  & 59.36  & 47.43  & 64.63  & 46.35  & 60.67  & 45.96  & 63.99  \\
\bottomrule
\end{tabular}
\caption{Performance of fine-grained consistency detection models in summaries generated by different systems (\%). 
``$-$ Doc. Fact Attention'' indicates that we remove the document fact attention module and use mean pooling to fuse all document semantic representation vectors. 
}
\label{table:model-type-results}
\vspace{-0.1in}
\end{table*}

\paragraph{Dataset. }
We conduct experiments on the Aggrefact-Unified dataset~\cite{DBLP:journals/corr/aggrefact21}, which
collects samples and unifies factual error types from four manually annotated datasets~\cite{DBLP:conf/acl/MaynezNBM20,DBLP:conf/naacl/Frank21,DBLP:conf/naacl/GoyalD21,DBLP:conf/emnlp/CLIFF21}. 
We remove the duplicated samples (i.e., duplicated document-summary pairs) in the Aggrefact-Unified dataset~\cite{DBLP:journals/corr/aggrefact21} and obtain 4,489 samples. We randomly split data samples into train/validation/test sets of size 3,689/300/500. 
The statistics of the error type labels are in Appendix~\ref{appendix:dataset-stat}.

\paragraph{Evaluation metrics. }
We adopt the macro-averaged F1 score and balanced accuracy (\textbf{BACC}) as the evaluation metrics. 
BACC is an extension of accuracy for class-imbalanced datasets and is widely adopted by previous literature on inconsistency detection~\cite{DBLP:conf/emnlp/Factcc20, DBLP:journals/tacl/summac22}. 
All experiment results are averaged across four random runs.

\paragraph{Baselines. }
We adapt the following baselines\footnote{We do not use QA-based metrics~\cite{DBLP:conf/emnlp/QuestEval21} as our baselines. It is because both noun phrase errors and predicate errors in the summary can cause a QA model to predict incorrect answers. Hence, we cannot decide the types of factual errors based on the outputs of QA-based metrics. } for the new task.  
\textbf{\textsc{FactCC-Multi}}: FactCC~\cite{DBLP:conf/emnlp/Factcc20} is originally trained on synthetic data for binary inconsistency detection. 
We replace the binary classifier with a multi-label classifier and finetune the model on Aggrefact. 
\textbf{\textsc{FactGraphMulti}}: FactGraph~\cite{DBLP:conf/naacl/factgraph22} parses each sentence into an AMR graph and uses a graph neural network to encode the document. We replace the binary classifier with a multi-label classifier. 
We also fine-tune the \textbf{\textsc{BERT}}~\cite{DBLP:conf/naacl/BERT19} and \textbf{\textsc{AdapterBert}}~\cite{DBLP:conf/icml/AdapterBERT19}.

\subsection{Performance of Error Type Detection}
Following \cite{DBLP:journals/corr/aggrefact21}, we detect error types in summaries from different models:
\textbf{SOTA} includes the pre-trained language models published in or after 2020. \textbf{XFORMER} contains the Transformer-based models published before 2020. \textbf{OLD} includes earlier RNN- or CNN-based models. \textbf{REF} represents reference summaries. 
From Table~\ref{table:model-type-results},
we observe that:
(1) \textit{Representing facts with semantic frames improves factual error type prediction.}. We observe that in most of the cases, our model outperforms other baselines that do not use semantic frames to represent facts. 
(2) \textit{The performance of our model drops after we remove the document fact attention module}. The results show that our document fact attention module \textit{not only improves the interpretability, but also boost the performance of factual error type detection}. 
(3) \textit{All detection models perform better in summaries generated by OLD systems}. It suggests that the factual errors made by OLD systems are relatively easier to recognize than the errors made by more advanced systems.

\begin{table}[t]
\centering
\small
\begin{tabular}{l c c c}
\toprule
\textbf{Model}       & \textbf{R@3}   & \textbf{R@4}   & \textbf{R@5} \\ \hline
\textsc{BERT} & 36.76 &  46.18 & 53.34 \\
\textsc{AdapterBert} & 36.34 &  46.14 & 53.80 \\
\textsc{FactCCMulti} & 41.11 & 50.95 & 58.41 \\
\textsc{FactGraphMulti} & 42.25 & 52.10 & 60.24 \\ \hdashline
\modelshort~ & \textbf{49.99} & \textbf{59.91} & \textbf{67.92} \\
\bottomrule
\end{tabular}
\caption{
The recall@3,4,5 scores of document fact highlights (\%). 
}
\vspace{-0.1in}
\label{table:fact-verify}
\end{table}

\subsection{Evaluation of Document Fact Highlights}
Since ground-truth document fact highlights are not available, we apply a fact verification dataset to evaluate whether the predicted document fact highlights provide evidence for inconsistency detection. 
Specifically, we adopt the FEVER 2.0 dataset~\cite{Thorne19FEVER2}, which consists of claims written by humans and evidence sentences from Wikipedia that can support or refute the claims. 
We first extract facts from the evidence sentences via SRL and use them as the \textit{ground-truth document fact highlights}. 
We then consider each claim as the input summary and the section of a Wikipedia article that contains the evidence sentences as the input document. 

We devise the following method to compute document fact highlights for the baseline models. 
Since all baselines utilize the CLS token to predict the factual error types, we use the attention scores received from the CLS token to compute an importance score for each document fact. 
We then return the facts that obtain the highest importance scores as the document fact highlights for each baseline. More details are in Appendix~\ref{appendix:baseline-fact-highlights}. 

Table~\ref{table:fact-verify} presents the recall scores of document fact highlights predicted by different models. We observe that our model obtains substantially higher recall scores, which demonstrates that our model provides more evidence to support the inconsistency prediction. Thus, compared with the baselines, our model allows users to verify the predicted error types and correct inconsistent summaries.

\begin{table}[t]
\small
\centering
\begin{tabular}{p{0.94\columnwidth}}
\toprule
\textbf{Source text:} \\ 
Children in P6 and P7 will learn how to cope with change under {the Healthy Me programme developed by Northern Ireland charity , Action Mental Health} 
...
The charity is now hoping the programme will be rolled out in schools across Northern Ireland ...
... 
\\
\hline
\textbf{Summary generated by an OLD model:} \\ 
\red{\textit{a school in northern ireland}} has launched a programme to help children with mental health problems in northern ireland .\\ \hline
\textbf{Ground-truth factual error type:} \\ 
Intrinsic Noun Phrase Error \\  \hline
\textbf{Factual error type predicted by \modelshort~:} \\ 
Intrinsic Noun Phrase Error \\ \hline
\textbf{Document fact highlight predicted by \modelshort~ ($k=1$):} \\ 
1. $[_{\text{ARG1}}$ the Healthy Me programme$]$ $[_{\text{V}}$ developed$]$ $[_{\text{ARG0}}$ by Northern Ireland charity , Action Mental Health$]$ \\
\bottomrule
\end{tabular}
\caption{
Sample outputs of our \modelshort~ model in the Aggrefact-Unified dataset. 
The error in the sample summary is in red color and italicized. 
}
\label{table:case-study}
\vspace{-0.1in}
\end{table}

\begin{table}[t]
\small
\centering
\begin{tabular}{p{0.94\columnwidth}}
\toprule
\textbf{Source text:} \\ 
The move is part of national fire service reforms unveiled by Home Secretary Theresa May last week . \textbf{Sussex PCC Katy Bourne} said emergency services would have an increased duty to collaborate under the new bill . But West Sussex County Council ( WSCC ) said it already had an excellent model . East Sussex ' s fire authority said it would co - operate with the PCC but it believed collaboration could be achieved without elaborate structural change . \textbf{Ms Bourne said she had written to WSCC leader Louise Goldsmith and Phil Howson , East Sussex Fire Authority chairman , to request they begin to look at the feasibility of bringing both fire services under her authority .}  ...
 \\
\hline
\textbf{Summary generated by a SOTA model:} \\ 
\textit{\red{West}} Sussex 's police and crime commissioner ( PCC ) has said she wants to look at the feasibility of bringing East Sussex 's fire service under her authority . \\ \hline
\textbf{Ground-truth factual error type:} \\ 
Intrinsic Noun Phrase Error \\  \hline
\textbf{Factual error type predicted by \modelshort~:} \\ 
No Error \\ \hline
\textbf{Document fact highlights predicted by \modelshort~ ($k=5$):} \\ 
1. $[_{\text{ARG1}}$ collaboration$]$ $[_{\text{ARGM-MOD}}$ could$]$ $[_{\text{V}}$ achieved$]$ $[_{\text{ARGM-MNR}}$ without elaborate structural change$]$ \\
2. $[_{\text{V}}$ bringing$]$ $[_{\text{ARG1}}$ both fire services$]$ $[_{\text{ARG3}}$ under her authority$]$ \\
3. $[_{\text{ARG0}}$ they$]$ $[_{\text{V}}$ begin$]$ $[_{\text{ARG1}}$ to look at the feasibility of bringing both fire services under her authority$]$ \\
4. $[_{\text{ARG0}}$ they$]$ $[_{\text{V}}$ look$]$ $[_{\text{ARG1}}$ at the feasibility of bringing both fire services under her authority$]$ \\
5. $[_{\text{ARG0}}$ she$]$ $[_{\text{V}}$ request$]$ $[_{\text{ARG1}}$ they begin to look at the feasibility of bringing both fire services under her authority$]$ \\
\bottomrule
\end{tabular}
\caption{
Incorrect output sample of our \modelshort~ model in the Aggrefact-Unified dataset~\cite{DBLP:journals/corr/aggrefact21}. 
The error in the sample summary is in red color and italicized. 
We bold the text spans from the document that refute the sample summary. 
}
\label{table:error-case}
\vspace{-0.1in}
\end{table}

\subsection{Case Study}
Table~\ref{table:case-study} shows a sample summary generated by an OLD model with an intrinsic noun phrase error, where the ``a school in northern ireland'' in the summary contradicts with ``Northern
Ireland charity" in the document.
Our model accurately predicts the error type with evidence in the form of document fact highlight, which helps users verify the error and correct the summary.

In Table~\ref{table:error-case}, we present an error analysis on a sample summary generated by a SOTA model. 
According to the source text, the word ``West'' in the summary is incorrect and should be removed since the statement in the summary is made by ``Sussex PPC'' instead of ``West Sussex PCC''. In order to detect this error, a model needs to understand that the expressions ``Sussex PCC Katy Bourne'', ``Ms Borune'', and ``she'' in the document refer to the same entity. This sample illustrates that the errors generated by a SOTA model are more subtle and more difficult to be detected. Our model fails to predict the correct error type for this sample. Since the top five document fact highlights returned by our model do not contain the entity ``Sussex PCC Katy Bourne'', we suspect that our model fails to recognize the co-referential relations among ``Sussex PCC Katy Bourne'', ``Ms Borune'', and ``she'' for this sample. Thus, improving the co-reference resolution ability of fine-grained inconsistency detection models is a potential future direction.

\section{Related Work}

\paragraph{Factual consistency metrics.} 
QA-based consistency metrics~\cite{DBLP:conf/acl/DurmusHD20,DBLP:conf/emnlp/QuestEval21,DBLP:conf/naacl/QAFactEval22} involve generating questions from the given document and its summary, and then comparing the corresponding answers to compute a factual consistency score. 
Entailment-based consistency metrics~\cite{DBLP:journals/tacl/summac22,DBLP:conf/emnlp/Factcc20,DBLP:conf/naacl/factgraph22,DBLP:conf/emnlp/DAE20} utilize a binary classifier to determine whether the contents in a system summary are entailed by the source article. 
In contrast, our model is a multi-label classifier that detects the types of factual errors in a summary. Moreover, our model leverages SRL to encode the facts in the input document and summary, enabling users to interpret which facts in the document are most relevant to the inconsistency detection. 

\paragraph{Fact-based evaluation methods.} 
To evaluate the informativeness of a summary, the Pyramid human evaluation protocol~\cite{DBLP:conf/naacl/Pyramid04} asks annotators to extract semantic content units (SCUs) from the system summary and reference summary, respectively, and then compute their overlap. Each SCU contains a single fact. 
\citet{DBLP:conf/acl/FactContentWeight20} approximate the Pyramid method by using SRL to extract facts. They then compute the embedding similarity between the facts extracted from the system summary and those from the reference summary. 
\citet{fischer2022measuring} also use SRL to extract facts, but they measure the similarity between the facts extracted from the system summary and those from the source document to compute a faithfulness score. 
On the other hand, our model integrates SRL with a multi-label classifier to predict the factual error types of a summary.

\section{Conclusion}
In this paper, we present a new task of fine-grained inconsistency detection, which aims to predict the types of factual inconsistencies in a summary. 
Compared to the previous binary inconsistency detection task, our new task can provide more insights into the weakness of summarization systems. 
Moreover, we propose an interpretable fine-grained inconsistency detection model, which represents facts from documents and summaries with semantic frames and highlights highly relevant document facts. 
Experiments on the Aggrefact-Unified dataset show that our model can better identify factual error types than strong baselines. Furthermore, results on the FEVER 2.0 dataset validate that the highlighted document facts provide evidence to support the inconsistency prediction.

\section{Limitations}
Although our model allows users to interpret which parts of the input document are most relevant to the model's prediction, our model does not allow users to interpret which text spans of the input summary contain errors. We use the summary in Table~\ref{table:case-study} as an example. If the model can indicate the text span ``a school in northern ireland'' contains errors, it will be easier for the user to correct the summary. 
\citet{DBLP:conf/emnlp/Factcc20} introduced an auxiliary task to extract erroneous text spans in summaries, but their method requires expensive text span ground-truth labels.
Locating incorrect text spans in the summaries without requiring span-level training labels remains unexplored. Another limitation of our model is that it does not allow users to interpret the uncertainty of the prediction results~\cite{DBLP:journals/tacl/DeutschDR21}.

\section{Ethical Considerations}
The factual error types and document fact highlights predicted by our model can help users correct factually inconsistent summaries. Since factually inconsistent summaries often convey misinformation, our model can potentially help users combat misinformation. However, the factual error types predicted by our model may be incorrect. For example, it is possible that an input summary contains extrinsic noun phrase errors, but our model predicts the error type of intrinsic predicate error. 
Hence, users still need to be cautious when using our model to detect and correct inconsistent summaries. The Aggrefact-Unified dataset contains public news articles from CNN, DailyMail, and BBC. Hence, the data that we used does not have privacy issues. 

\section*{Acknowledgement}
We thank the anonymous reviewers for their insightful comments on our work. This research is based upon work supported by U.S. DARPA AIDA Program No. FA8750-18-2-0014, DARPA INCAS Program No. HR001121C0165, NSF under award No. 2034562, the Molecule Maker Lab Institute: an AI research institute program supported by NSF under award No. 2019897 and No. 2034562, and the AI Research Institutes program by National Science Foundation and the Institute of Education Sciences, U.S. Department of Education through Award \# 2229873 - AI Institute for Transforming Education for Children with Speech and Language Processing Challenges. 
The views and conclusions contained herein are those of the authors and should not be interpreted as necessarily representing the official policies, either expressed or implied, of the U.S. Government, the National Science Foundation, the Institute of Education Sciences, or the U.S. Department of Education. The U.S. Government is authorized to reproduce and distribute reprints for governmental purposes notwithstanding any copyright annotation therein. 
Hou Pong Chan was supported in part by the Science and Technology Development Fund, Macau SAR (Grant Nos. FDCT/060/2022/AFJ, FDCT/0070/2022/AMJ) and the Multi-year Research Grant from the University of Macau (Grant No. MYRG2020-00054-FST).

\bibliography{anthology,custom}
\bibliographystyle{acl_natbib}

\appendix
\clearpage

\section{Details of Training Objective}\label{appendix:loss-function}
Since some error types may have an imbalanced distribution of positive and negative samples, we apply sampling weighting to the training objective. We first weigh the loss for the positive samples according to their proportion in the training set. Then we sum up the binary cross-entropy loss of each error type as the training objective. The weighted binary cross-entropy (BCE) loss of our model is formally defined as follows:
\begin{align}
    L_{i} &= \beta_{i} y_{i}^{*} \log p(y_{i}) + (1 - y_{i}^{*}) \log (1-p(y_{i}))\text{,} \\
    L &= \sum_{i=1}^{K}  L_{i}\text{,}
\end{align}
where $\beta_{i}$ is the weight for positive samples of the $i$-th error type. We set $\beta_{i}$ to be the ratio of the number of positive samples to the number of negative samples of the $i$-th error type in the training data.

\section{Experiment Details}
\subsection{Aggrefact-Unified Dataset}\label{appendix:dataset-stat}
This dataset contains news documents from CNN/DM~\cite{DBLP:conf/conll/NallapatiZSGX16} and XSum~\cite{DBLP:conf/emnlp/XSum18}. 
In addition to the four factual error types presented in Table~\ref{table:error-types-examples}, the Aggrefact-Unified dataset also provides the labels of \textit{intrinsic entire-sentence error}, \textit{extrinsic entire-sentence error}, and \textit{entire-sentence error}. We map intrinsic (extrinsic) entire-sentence errors to intrinsic (extrinsic) noun phrases and intrinsic (extrinsic) predicate errors. We also map the entire-sentence error to all four types of factual errors. 
Statistics of the factual error type labels are shown in Table~\ref{table:error-types-stat}.  Table~\ref{table:model-types-stat} presents the statistics of summaries generated by different systems. 

\subsection{Extraction of Document Fact Highlights for Baseline Models}\label{appendix:baseline-fact-highlights}
Given a baseline model and a sample output from the baseline model, we first extract all the facts from the input document by SRL. 
Then for each extracted document fact, we compute the average attention score injected from the CLS token to the tokens in the semantic frame in the last layer of the baseline model. This average attention score is treated as the importance score of the document fact. 
Concretely, we use $\alpha'_{\text{CLS} \rightarrow i }$ to denote the total attention score injected from the CLS token to the $i$-th token of the semantic frame in the last layer of the baseline model over all attention heads. Then we compute the importance score as follows: $\sum_{i=1}^{m} \alpha'_{\text{CLS} \rightarrow i}$, where $m$ is the number of words in the fact. Finally, we return the document facts with the highest importance scores as the document fact highlights.

\subsection{Hyper-parameter Settings}
To compute F1 and BACC scores, we set the classification threshold to be 0.5. 
The dimension of the adapter in the Adapter-BERT model is set to 32. 
The number of attention heads in our document fact attention module is set to 16. 
We search the optimal number of attention heads from $\{1,4,8,16\}$ that obtains the highest BACC score in the validation set. We train our models for 40 epochs and select the checkpoint that obtains the highest BACC score in the validation set. We set the learning rate to be 1e-5. The training batch size is 12 with a gradient accumulation steps of 2. 
The AdapterBERT, BERT, and FineGrainFact models receive the same amount of hyperparameter tuning. 

\begin{table}[t]
\small
\centering
\begin{tabular}{l c c c c c c}
\toprule
\textbf{Source} & \textbf{Ex. NP} & \textbf{In. NP} & \textbf{Ex. Pred.} & \textbf{In. Pred.}  \\ \hline
CNNDM & 348 & 200 & 280 & 111 \\
XSum & 1,812 & 1,114 & 540 & 327  \\
\bottomrule
\end{tabular}
\caption{Statistics of fine-grained error types in the AggreFact-Unified dataset. }
\label{table:error-types-stat}
\end{table}

\begin{table}[t]
\small
\centering
\begin{tabular}{l c c c c}
\toprule
\textbf{Source} & \textbf{SOTA} & \textbf{XFORMER} & \textbf{OLD} & \textbf{REF}  \\ \hline
CNNDM & 550 & 249 & 800 & 0 \\
XSum & 400 & 994 & 997 & 499  \\
\bottomrule
\end{tabular}
\caption{Statistics of summaries generated by different systems in the AggreFact-Unified dataset. }
\label{table:model-types-stat}
\end{table}

\subsection{Hardware and Software Configurations}
We run all the experiments using a single NVIDIA V100 GPU. It takes around 1 hour and 50 minutes to train our model for 40 epochs. Our model contains 113.1M of parameters in total. We only need to train 3.6M of the model parameters since most of the parameters are frozen by the Adapter-BERT model. We obtain the BERT-base-uncased checkpoint from Huggingface~\cite{DBLP:journals/corr/huggingface19}. We adopt the implementation of the BERT-based SRL model~\cite{DBLP:journals/corr/SRL19} provided by AllenNLP~\cite{allennlp18} to conduct semantic role labeling~\cite{DBLP:journals/coling/SRL05}.

\section{Results on Different Summarization Datasets and Error Types}
In Table~\ref{table:separate-dataset-and-error-types}, we separate the F1 scores obtained by our \modelshort~ model according to the summarization dataset and the type of factual errors. It is observed that our model has relatively low performance ($<50 \%$) on detecting intrinsic errors (intrinsic noun phrase and intrinsic predicate errors) in the XSum dataset. We analyze the reason as follows. According to previous studies~\cite{DBLP:conf/acl/DurmusHD20}, system summaries generated in the XSum dataset tend to have a high abstractiveness (low textual overlapping with the source document). We suspect that our \modelshort~ model learns a spurious correlation that suggests an inconsistent summary with high abstractiveness contains extrinsic errors rather than intrinsic errors. A critical future direction is to address this spurious correlation of our model.

\begin{table}[t]
\centering
\small
\begin{tabular}{l c c}
\toprule
\textbf{Error Type}       & \textbf{XSum}   & \textbf{CNN/DM}  \\ \hline
Extrinsic NP & 64.58 &  52.39 \\
Extrinsic Pred. & 64.26 &  52.15 \\
Intrinsic NP & 46.48 &  63.01 \\
Intrinsic Pred. & 42.61 &  51.53 \\
\bottomrule
\end{tabular}
\caption{
The F1 score results of the \modelshort~ model in each summarization dataset and factual error type (\%). 
}
\vspace{-0.1in}
\label{table:separate-dataset-and-error-types}
\end{table}

\section{Generalization Ability Analysis}
To more robustly evaluate the generalization ability of inconsistency detection models, we further construct a challenging data split in which there are no overlapped systems and documents between the test set and the training set. We first gather all the samples that contain a summary generated by the BART model~\cite{DBLP:conf/acl/BART20} to construct the test set. We choose BART since it is a common baseline in recent summarization literature~\cite{DBLP:journals/corr/sumren,DBLP:conf/aaai/ZhongLX0022}. After that, we randomly split the remaining data samples into training and validation sets. Finally, we remove the duplicated documents between the training set and the test set. This data split contains 3,839/550/100 samples for train/validation/test sets. The results of different inconsistency detection models are shown in Table~\ref{table:generalization-ability-analysis}. We observe that our \modelshort~ model outperforms all the baselines, which demonstrates the strong generalization ability of our model. 

\begin{table}[t]
\centering
\small
\begin{tabular}{l c c}
\toprule
\textbf{Model}       & \textbf{F1}   & \textbf{BACC}  \\ \hline
\textsc{BERT} & 38.83 &  59.27 \\
\textsc{AdapterBert} & 39.88 &  61.20 \\
\textsc{FactCCMulti} & 32.53 & 58.24 \\
\textsc{FactGraphMulti} & 25.83 & 57.55 \\ \hdashline
\modelshort~ & \textbf{40.71} & \textbf{62.19} \\
\bottomrule
\end{tabular}
\caption{
Performance of fine-grained inconsistency detection models in the challenging data split (\%). 
}
\vspace{-0.1in}
\label{table:generalization-ability-analysis}
\end{table}

\section{Scientific Artifacts}
We list the licenses of the scientific artifacts used in this paper: AllenNLP (Apache License 2.0), Huggingface Transformers (Apache License 2.0), and \textsc{FactCC} (BSD-3-Clause License). We apply the above artifacts according to their official documentation. We will release an API of our model for research purposes. Our API can be applied to detect the fine-grained factual error types in summaries written in the English language.

\end{document}